\documentclass[11pt,a4paper]{article}
\usepackage{times,latexsym}
\usepackage{url}
\usepackage[T1]{fontenc}

\usepackage[acceptedWithA]{tacl2018v2}
\usepackage{microtype}
\usepackage{amsmath}
\usepackage{amssymb}
\usepackage{multirow}
\usepackage{graphicx}
\usepackage{subcaption}
\usepackage{ifthen}
\usepackage{xspace}
\usepackage{xcolor}
\usepackage{booktabs}
\usepackage{colortbl}
\usepackage{tabularx}
\usepackage{comment}

\newcommand\refsec[1]{Section~\ref{sec:#1}}

\newcommand\reffig[1]{Figure~\ref{fig:#1}}

\newcommand\reftab[1]{Table~\ref{tab:#1}}

\ifthenelse{\isundefined{\definition}}{\newtheorem{definition}{Definition}}{}
\ifthenelse{\isundefined{\assumption}}{\newtheorem{assumption}{Assumption}}{}
\ifthenelse{\isundefined{\hypothesis}}{\newtheorem{hypothesis}{Hypothesis}}{}
\ifthenelse{\isundefined{\proposition}}{\newtheorem{proposition}{Proposition}}{}
\ifthenelse{\isundefined{\theorem}}{\newtheorem{theorem}{Theorem}}{}
\ifthenelse{\isundefined{\lemma}}{\newtheorem{lemma}{Lemma}}{}
\ifthenelse{\isundefined{\corollary}}{\newtheorem{corollary}{Corollary}}{}
\ifthenelse{\isundefined{\alg}}{\newtheorem{alg}{Algorithm}}{}
\ifthenelse{\isundefined{\example}}{\newtheorem{example}{Example}}{}

\newcommand\nl[1]{``\textit{#1}''}
\newcommand\eg{e.g.,\xspace}
\newcommand\ie{i.e.\xspace}

\newcolumntype{L}[1]{>{\raggedright\let\newline\\\arraybackslash\hspace{0pt}}m{#1}}
\newcolumntype{C}[1]{>{\centering\let\newline\\\arraybackslash\hspace{0pt}}m{#1}}
\newcolumntype{R}[1]{>{\raggedleft\let\newline\\\arraybackslash\hspace{0pt}}m{#1}}
\newcommand{\cent}[1]{\multicolumn{1}{c}{#1}}
\newcommand{\std}{\hphantom{ (0.0)}\xspace}

\newcommand{\bert}{BERT$_{\text{BASE}}$\xspace}

\newcommand{\roberta}{RoBERTa$_{\text{BASE}}$\xspace}
\newcommand{\bertl}{BERT$_{\text{LARGE}}$\xspace}
\newcommand{\robertal}{RoBERTa$_{\text{LARGE}}$\xspace}
\newcommand{\rbert}{BERT$_{\text{scratch}}$\xspace}
\newcommand{\hans}{HANS\xspace}
\newcommand{\paws}{PAWS\xspace}
\newcommand{\mnli}{MNLI\xspace}

\newcommand{\qqp}{QQP\xspace}
\newcommand{\pawsqqp}{PAWS$_{\text{QQP}}$\xspace}
\newcommand{\pawswiki}{PAWS$_{\text{Wiki}}$\xspace}
\newcommand{\pawsall}{PAWS$_{\text{QQP+Wiki}}$\xspace}

\title{An Empirical Study on Robustness to Spurious Correlations using Pre-trained Language Models}

\author{
 Lifu Tu$^1$ \Thanks{Most work was done during first author's internship and last author's work at Amazon AI.}\ \ \ \ \ \ Garima Lalwani$^2$ \ \ \  \ \ \  Spandana Gella$^2$ \ \ \ \ \ \ He He$^3$ \footnotemark[1]  \\
 $^1$Toyota Technological Institute at Chicago, 
 $^2$Amazon AI, \ \ \ $^3$New York University \\
  {lifu@ttic.edu, \{glalwani, sgella\}@amazon.com, hehe@cs.nyu.edu} \\
}

\date{}

\begin{document}
\maketitle
\begin{abstract}
    Recent work has shown that pre-trained language models such as BERT
    improve robustness to spurious correlations in the dataset. 
    Intrigued by these results, 
    we find that the key to their success is
    generalization from a small amount of counterexamples
    where the spurious correlations do not hold.
    When such minority examples are scarce, pre-trained models perform as poorly as models trained from scratch. 
    In the case of extreme minority,
    we propose to use multi-task learning (MTL) to improve generalization.
    Our experiments on natural language inference and paraphrase identification show that
    MTL with the right auxiliary tasks significantly improves performance on challenging examples without hurting the in-distribution performance.
    Further, we show that the gain from MTL mainly comes from improved generalization from the minority examples.
    Our results highlight the importance of data diversity for overcoming spurious correlations.\footnote{Code 
is available at \url{https://github.com/lifu-tu/Study-NLP-Robustness}}
\end{abstract}

\section{Introduction}
\label{introduction}
A key challenge in building robust NLP models is the gap between limited linguistic variations in the training data and the diversity in real-world languages.
Thus models trained on a specific dataset are likely to rely on \emph{spurious correlations}:
prediction rules that work for the majority examples but do not hold in general. 
For example, in natural language inference (NLI) tasks, previous work has found that
models learned on notable benchmarks achieve high accuracy
by associating high word overlap between the premise and the hypothesis with entailment \citep{DasguptaGSGG18,mccoy2019hans}.
Consequently, these models perform poorly on the so-called \emph{challenging} or \emph{adversarial datasets} where such correlations no longer hold \citep{glockner2018breaking,mccoy2019hans,nie2019analyzing,zhang2019paws}. 
This issue has also been referred to as annotation artifacts \citep{gururangan2018annotation}, dataset bias \citep{he2019unlearn,clark2019ensemble}, and group shift \citep{oren2019distributionally,sagawa2020distributionally} in the literature.

Most current methods rely on prior knowledge of spurious correlations in the dataset
and tend to suffer from a trade-off between \emph{in-distribution accuracy} on the independent and identically distributed (i.i.d.) test set and
\emph{robust accuracy\footnote{We use the term ``robust accuracy'' from now on to refer to the accuracy on challenging datasets.}} on the challenging dataset.
Nevertheless, recent empirical results have suggested that self-supervised pre-training improves robust accuracy, 
while not using any task-specific knowledge nor incurring in-distribution accuracy drop \citep{hendrycks2019pretraining,hendrycks2020pretrained}. 

\begin{table*}[ht]
    \centering
    \definecolor{Gray}{gray}{0.9}
    \footnotesize{
        \begin{tabular}{llrllll}
        \toprule
            & Dataset & Size & Heuristic & Input & Label \\
        \midrule
            \multicolumn{6}{c}{Natural language inference} \\
        \midrule
            \multirow{2}{*}{Train} & \multirow{2}{*}{MNLI} & \multirow{2}{*}{393k}
                & high word overlap & P: {\color{blue}The doctor mentioned the manager} who ran. & \multirow{2}{*}{entailment}\\
            &
            &  & $\Rightarrow$ entailment & H: {\color{blue}The doctor mentioned the manager}. & \\
            \rowcolor{Gray}
            && 
                & high word overlap & P: The actors who advised {\color{blue}the manager saw the tourists}. & \\
            \rowcolor{Gray}
            \multirow{-2}{*}{Test} & \multirow{-2}{*}{\hans} & \multirow{-2}{*}{30k}
               & $\nRightarrow$ entailment & H: {\color{blue}The manager saw the tourists}. & \multirow{-2}{*}{non-entailment}\\
        \midrule
            \multicolumn{6}{c}{Paraphrase Identification} \\
        \midrule
            \multirow{2}{*}{Train} & \multirow{2}{*}{QQP} & \multirow{2}{*}{364k}
                & same bag-of-words & S$_1$: {\color{red}Bangkok} vs {\color{blue}Shanghai}? & \multirow{2}{*}{paraphrase}\\
            &
            &  & $\Rightarrow$ paraphrase & S$_2$: {\color{blue}Shanghai} vs {\color{red}Bangkok}? & \\
            \rowcolor{Gray}
            && 
                & same bag-of-words & S$_1$: Are all dogs {\color{red}smart} or can some be {\color{blue}dumb}? & \\
            \rowcolor{Gray}
            \multirow{-2}{*}{Test} & \multirow{-2}{*}{\pawsqqp} & \multirow{-2}{*}{677}
              & $\nRightarrow$ paraphrase & S$_2$: Are all dogs {\color{blue}dumb} or can some be {\color{red}smart}? & \multirow{-2}{*}{non-paraphrase}\\
        \bottomrule
    \end{tabular}
    }
    \caption{\label{tab:dataset-examples}
    Representative examples from the training datasets (MNLI and QQP) and the challenging/test datasets (\hans and \pawsqqp).
    Overlaping text spans are highlighted for NLI examples
    and swapped words are highlighted for paraphrase identification examples.
    The word overlap-based heuristic that works for typical training examples fails on the test data.
    }
\end{table*}

\begin{table*}[ht]
  \begin{center}
    \begin{tabular}{lllll} 
    \toprule
              & \multicolumn{2}{c}{\textit{Trained on MNLI}} & \multicolumn{2}{c}{\textit{Trained on QQP}} \\
              \cmidrule(lr){2-3} \cmidrule(lr){4-5}
              & \cent{In-distribution} & \cent{Challenging} & \cent{In-distribution} & \cent{Challenging} \\
        Model  & \cent{\textbf{MNLI-m}} & \cent{\textbf{HANS}} & \cent{\textbf{QQP}} & \cent{\textbf{\pawsqqp}}\\
      \midrule
      \multicolumn{3}{l}{\textit{Non pre-trained baselines}} \\
      \rbert  & 67.9 (0.5) & 49.9 (0.2) & 83.0 (0.7) & 40.6 (1.9) \\
        ESIM  & 78.1$^a$\std  & 49.1$^a$\std  & 85.3$^b$\std  & 38.9$^b$\std \\
      \midrule
      \multicolumn{3}{l}{\textit{pre-trained models}} \\
      \bert(prior)     & 84.0$^c$   & 53.8$^c$    & 90.5$^d$   & 33.5$^d$   \\
      \bert(ours)     & 84.5 (0.1) & 62.5 (3.4)  & 90.8 (0.3) & 36.1 (0.8) \\
      \bertl    & 86.2 (0.2) & 71.4 (0.6)  & 91.3 (0.3) & 40.1 (1.8) \\
      \roberta  & 87.4 (0.2) & 74.1 (0.9)  & 91.5 (0.2) & 42.6 (1.9) \\
      \robertal & 89.1 (0.1) & 77.1 (1.6)  & 89.0 (3.1) & 39.5 (4.8) \\
      \bottomrule
    \end{tabular}
      \caption{\label{tab:bert-finetuning}Accuracies (with standard deviation) on the in-distribution datasets, MNLI-matched (MNLI-m) and QQP dev sets,
      as well as the challenging datasets, \hans and \pawsqqp.
      Pre-trained transformers improve accuracies on both the in-distribution and challenging datasets over non pre-trained models,
      except on \pawsqqp.
      Our models fine-tuned for more epochs further improve prior results on the challenging data.
      Results taken from prior work:
      $^a$ \citet{he2019unlearn}, $^b$ \citet{zhang2019paws}, $^c$ \citet{mccoy2019hans}, $^d$ \citet{zhang2019paws}.
  }
  \end{center}
\end{table*}

In this paper,  
we aim to investigate how and when pre-trained language models such as BERT 
improve performance on challenging datasets.
Our key finding is that pre-trained models are more robust to spurious correlations
because they can generalize from a minority of training examples that \emph{counter} the spurious pattern,
\eg non-entailment examples with high premise-hypothesis word overlap.
Specifically, removing these counterexamples from the training set significantly hurts their performance on the challenging datasets.
In addition, larger model size, more pre-training data, and longer fine-tuning further improve robust accuracy.
Nevertheless, pre-trained models still suffer from spurious correlations when there are too few counterexamples.
In the case of extreme minority, we empirically show that multi-task learning (MTL) improves robust accuracy by 
improving generalization from the minority examples,
even though preivous work has suggested that MTL has limited advantage in i.i.d. settings \cite{sogaard2016deep,hashimoto2017joint}.

This work sheds light on the effectiveness of pre-training on robustness to spurious correlations.
Our results highlight the importance of data diversity (even if the variations are imbalanced).
The improvement from MTL also suggests that
traditional techniques that improve generalization in the 
i.i.d. 
setting 
can also improve out-of-distribution generalization
through the minority examples.

\section{Challenging Datasets}

In a typical supervised learning setting,
we test the model on held-out examples drawn from the same distribution as the training data,
\ie the in-distribution or i.i.d. 
test set.
To evaluate if the model latches onto known spurious correlations,
challenging examples are drawn from a different distribution where such correlations do not hold.
In practice, these examples are usually adapted from the in-distribution examples to counter known spurious correlations on notable benchmarks.
Poor performance on the challenging dataset is considered an indicator of a problematic model that relies on spurious correlations between inputs and labels.
Our goal is to develop robust models that have good performance on both the i.i.d. 
test set and the challenging test set.

\subsection{Datasets}
\label{sec:datasets}
We focus on two natural language understanding tasks, NLI and paraphrase identification (PI).
Both have large-scale benchmarking datasets with around 400k examples.
While recent models have achieved near-human performance on these benchmarks,\footnote{See the leaderboard at \url{https://gluebenchmark.com}.
}
the challenging datasets exploiting spurious correlations bring down the performance of state-of-the-art models below random guessing.
We summarize the datasets used for our analysis in \reftab{dataset-examples}.

\paragraph{NLI.}
Given a premise sentence and a hypothesis sentence, the task is to predict whether the hypothesis is entailed by, neutral with, or contradicts the premise.
MultiNLI (MNLI) \citep{williams2017broad} is the most widely used benchmark for NLI, and it is also the most thoroughly studied in terms of spurious correlations.
It was collected using the same crowdsourcing protocol as its predecessor SNLI \citep{bowman2015large} but covers more domains.
Recently, \citet{mccoy2019hans} exploit
high word overlap between the premise and the hypothesis for entailment examples to construct a challenging dataset called \hans.
They use syntactic rules to generate non-entailment (neutral or contradicting) examples with high premise-hypothesis overlap.
The dataset is further split into three categories depending on the rules used:
\texttt{lexical overlap}, \texttt{subsequence}, and \texttt{constituent}.

\paragraph{PI.}
Given two sentences, the task is to predict whether they are paraphrases or not.
On Quora Question Pairs (QQP) \citep{iyer2017qqp}, one of the largest PI dataset,
\citet{zhang2019paws} show that very few non-paraphrase pairs have high word overlap. 
They then created a challenging dataset called \paws that contains sentence pairs with high word overlap 
but different meanings
through word swapping and back-translation.
In addition to \pawsqqp which is created from sentences in QQP,
they also released \pawswiki created from Wikipedia sentences.

\section{Pre-training Improve Robust Accuracy}
\label{sec:finetuning}
Recent results have shown that pre-trained models appear to improve performance on challenging examples over models trained from scratch \cite{yaghoobzadeh2019,he2019unlearn,kaushik2020learning}.
In this section, we confirm this observation by thorough experiments on different pre-trained models
and motivate our inquiries.

\paragraph{Models.}
We compare pre-trained models of different sizes and using different amounts of pre-training data.
Specifically, we use the \bert (110M parameters) and \bertl (340M parameters) models implemented in GluonNLP \citep{guo2020gluoncv}
pre-trained on 16GB of text 
\citep{devlin2019BERT}.\footnote{
    The \texttt{book\_corpus\_wiki\_en\_uncased} model from \url{https://gluon-nlp.mxnet.io/model\_zoo/bert/index.html}.
}
To investigate the effect of size of the pre-training data, we also experiment with the \roberta and \robertal models \citep{liu2019roberta},\footnote{
    The \texttt{openwebtext\_ccnews\_stories\_books\_cased} model from \url{https://gluon-nlp.mxnet.io/model\_zoo/bert/index.html}.
}
which have the same architecture as BERT but were trained on ten times as much text (about 160GB).
To ablate the effect of pre-training, we also include a \bert model with random initialization, \rbert.

\paragraph{Fine-tuning.}
We fine-tuned all models for 20 epochs and selected the best model based on the in-distribution dev set. 
We used the Adam optimizer with a learning rate of 2e-5,
L2 weight decay of 0.01,
batch sizes of 32 and 16 for \texttt{base} and \texttt{large} models respectively. 
Weights of \rbert and the last layer (classifier) of pre-trained models are initialized from 
a normal distribution with zero mean and 0.02 variance.
All experiments are run with 5 random seeds and the average values are reported.

\begin{figure*}[ht]
  \centering
  \begin{subfigure}[b]{0.49\linewidth}
    \includegraphics[width=\linewidth]{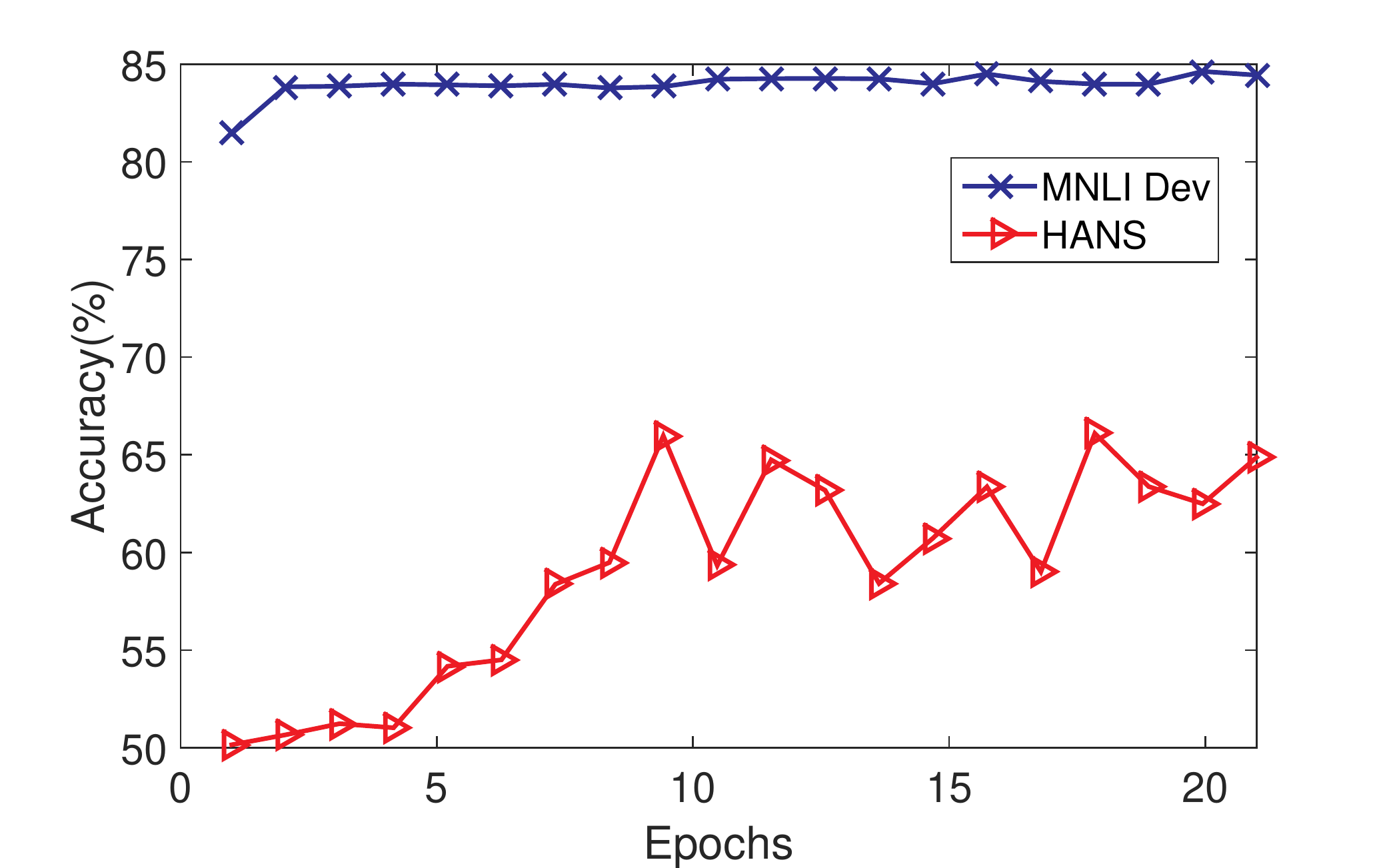}
    \caption{NLI/HANS}
  \end{subfigure}
  \begin{subfigure}[b]{0.49\linewidth}
    \includegraphics[width=\linewidth]{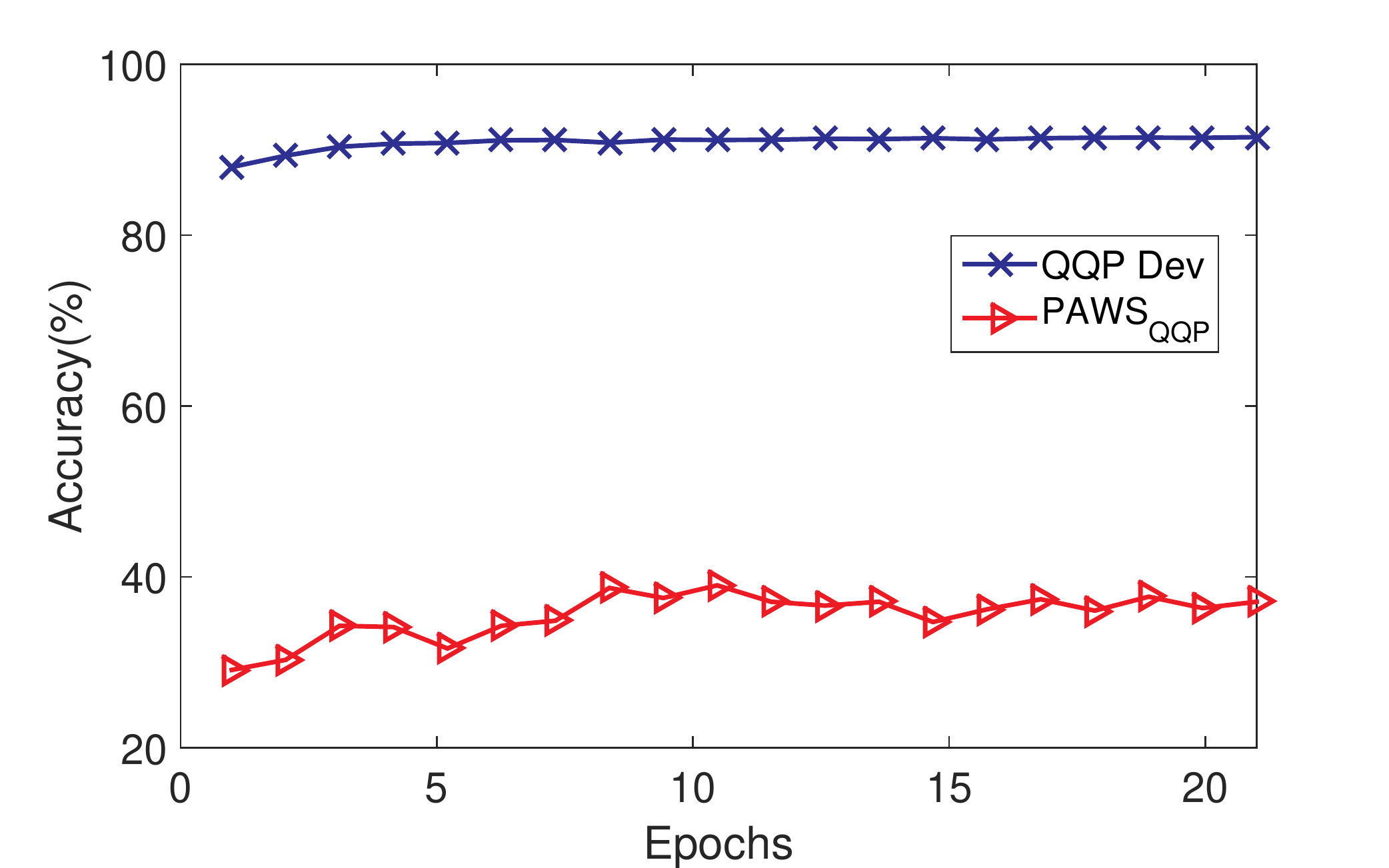}
    \caption{QQP/PAWS}
  \end{subfigure}
    \caption{Accuracy on the in-distribution data (MNLI dev and QQP dev)
    and the challenging data (HANS and \pawsqqp)
    after each fine-tuning epoch using \bert.
    The performance plateaus on the in-distribution data quickly around epoch 3,
    however, accuracy on the challenging data keeps increasing.
  }
  \label{fig:bert-finetuning}
\end{figure*}

\paragraph{Observations and inquiries.}
In Table~\ref{tab:bert-finetuning}, we show results for NLI and PI respectively.
As expected, they improve performance on in-distribution test sets significantly.\footnote{
    The lower performance of \robertal compared to \roberta is partly due to
    its high variance in our experiments.
}
On the challenging datasets, we make two key observations.

First, while pre-trained models improve the performance on challenging datasets,
the improvement is not consistent across datasets.
Specifically, 
the improvement on \pawsqqp are less promising than \hans.
While larger models (\texttt{large} vs. \texttt{base}) 
and more training data (RoBERTa vs. BERT)
yield a further improvement of 5 to 10 accuracy points on \hans,
the improvement on \pawsqqp is marginal.

Second,
even though three to four epochs of fine-tuning is typically sufficient for in-distribution data, 
we observe that longer fine-tuning improves results on challenging examples significantly
(see \bert ours vs. prior in \reftab{bert-finetuning}).
As shown in \reffig{bert-finetuning}, 
while the accuracy on MNLI and QQP dev sets
saturate after three epochs,
the performance on the corresponding challenging datasets keeps increasing until around the tenth epoch,
with more than 30\% improvement.

The above observations motivate us to ask the following questions:
\begin{enumerate}
    \item  How do pre-trained models generalize to out-of-distribution data?
    \item  When do they generalize well given the inconsistent improvements?
    \item  What role does longer fine-tuning play?
\end{enumerate}
We provide empirical answers to these questions in the next section
and show that the answers are all related to a small amount of counterexamples in the training data.

\begin{figure*}[ht]
    \centering
    \includegraphics[width=\linewidth]{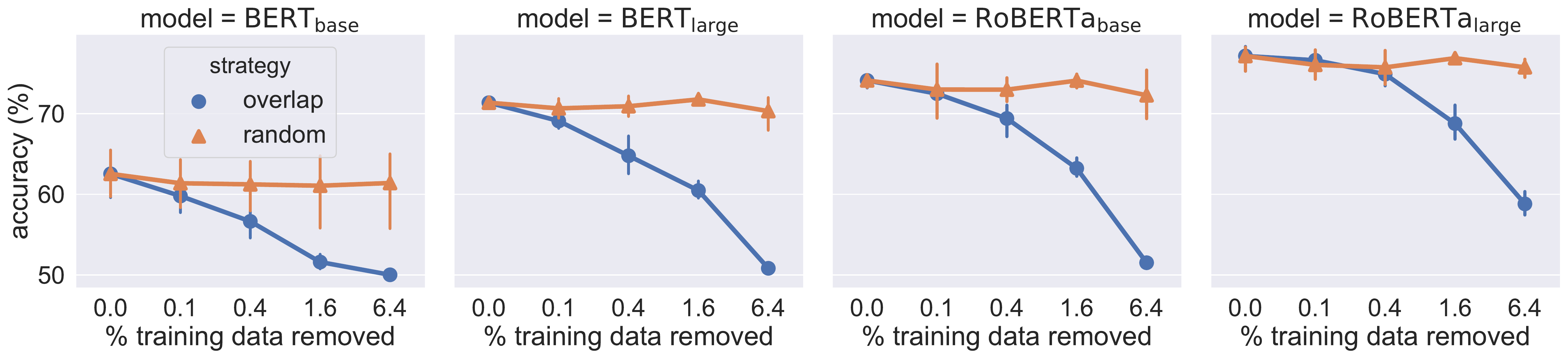}
    \caption{Accuracy on \hans when a small fraction of training data is removed. Removing non-entailment examples with high premise-hypothesis overlap significantly hurt performance compared to removing examples uniformly at random.}
    \label{fig:remove}
\end{figure*}

\section{Generalization from Minority Examples}
\subsection{Pre-training Improves Robustness to Data Imbalance}
\label{sec:pretraining-robustness}
One common impression is that the diversity in large amounts of pre-training data
allows pretrained models to generalize better to out-of-distribution data.
Here we show that while pre-training improves generalization, they do not enable \emph{extrapolation} to unseen patterns.
Instead, they generalize better from minority patterns in the training set.

Importantly, we notice that examples in \hans and \paws are not completely uncovered by the training data,
but belong to the minority groups.\footnote{
    Following \citet{sagawa2020distributionally}, we loosely define \emph{group} as a distribution of examples with similar patterns, \eg high premise-hypothesis overlap and non-entailment.
}
For example, in \mnli, there are 727 HANS-like non-entailment examples where all words in the hypothesis also occur in the premise;
in \qqp, there are 247 PAWS-like non-paraphrase examples where the two sentences have the same bag of words.
We refer to these examples that counter the spurious correlations as \emph{minority examples}.
We hypothesize that pre-trained models are more robust to 
group imbalance,
thus generalizing well from the minority groups.

To verify our hypothesis,
we remove minority examples during training
and observe its effect on robust accuracy.
Specifically, for NLI we sort non-entailment (contradiction and neutral) examples in MNLI by their premise-hypothesis overlap,
which is defined as the percentage of hypothesis words that also appear in the premise.
We then remove increasing amounts of these examples in the sorted order. 

As shown in \reffig{remove},
all models have significantly worse accuracy on \hans as more counterexamples are removed,
while maintaining the original accuracy when the same amounts of random training examples are removed.
With 6.4\% counterexamples removed, the performance of most pretrained models is near-random,
as poor as non-pretrained models.
Interestingly, larger models with more pre-training data (\robertal)
appear to be slightly more robust with increased level of imbalance.

\paragraph{Takeaway.}
These results reveal that pre-training improve robust accuracy by improving the i.i.d. accuracy on minority groups,
highlighting the importance of increasing data diversity when creating benchmarks.
Further, pre-trained models still suffer from suprious correlations
when the minority examples are scarce.
To enable extrapolation, we might need additional inductive bias \citep{nye2019learning} or new learning algorithms \citep{arjovsky2019invariant}.

\subsection{Minority Patterns Require Varying Amounts of Training Data}
\label{sec:paws}
Given that pre-trained models generalize better from minority examples,
why do we not see similar improvement on \pawsqqp
even though QQP also contains counterexamples?
Unlike \hans examples that are generated from a handful of templates,
\paws examples are generated by swapping words 
in a sentence followed by human inspection.
They often require recognizing nuance syntactic differences between two sentences with a small edit distance.
For example, compare
\nl{What's classy if you're poor , but trashy if you're rich?} and \nl{What's classy if you're rich , but trashy if you're poor?}.
Therefore, we posit that more samples are needed to reach good performance on \paws-like examples.

\begin{figure}[ht]
    \centering
    \includegraphics[width=\linewidth]{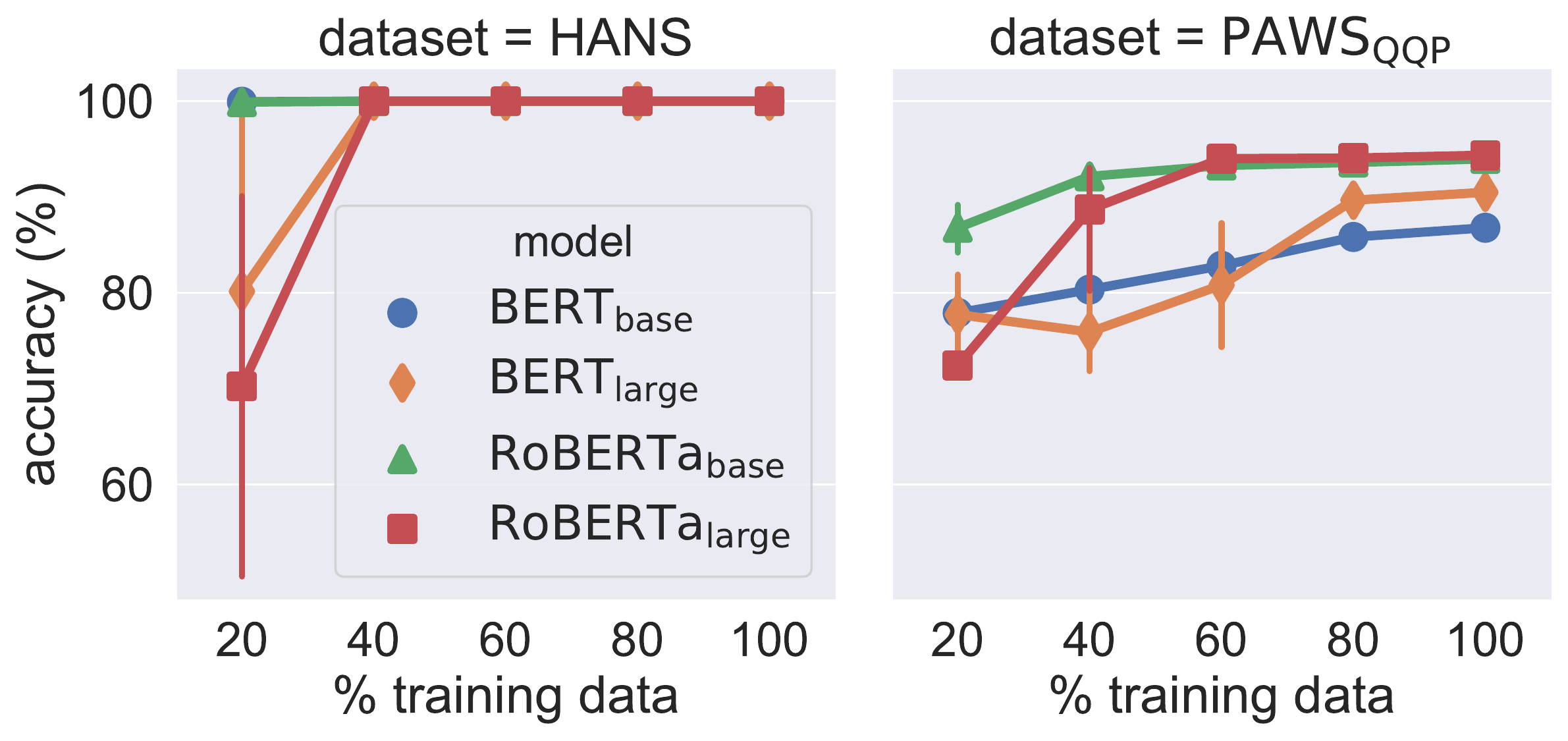}
    \caption{Learning curves of models trained on \hans and \pawsqqp.
    Accuracy on \pawsqqp increases slowly,
    whereas all models quickly reach 100\% accuracy on \hans.}
    \label{fig:learning_curve}
\end{figure}

\begin{figure*}[ht]
  \centering
  \begin{subfigure}[b]{0.49\linewidth}
    \includegraphics[width=\linewidth]{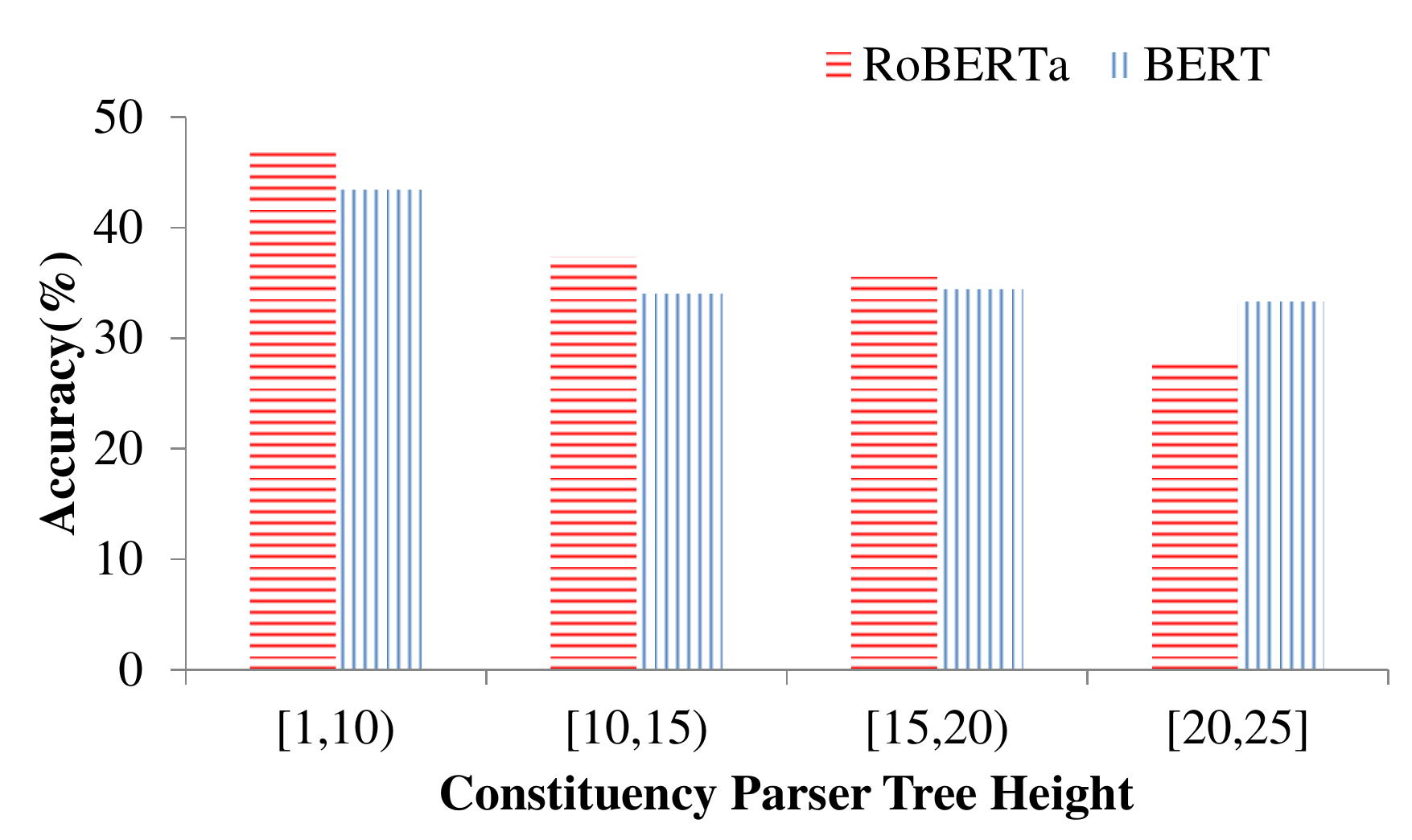}
  \end{subfigure}
  \begin{subfigure}[b]{0.49\linewidth}
    \includegraphics[width=\linewidth]{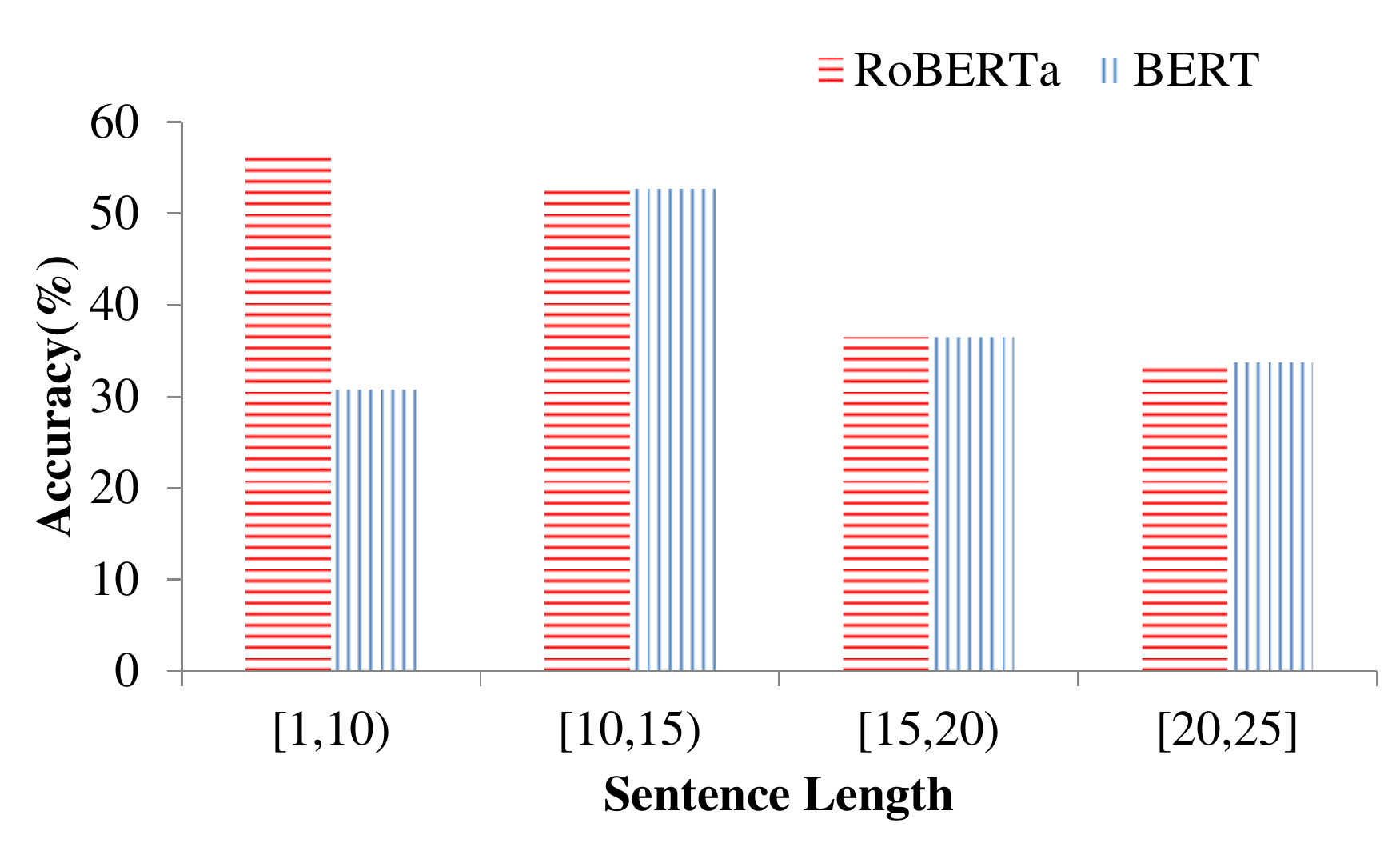}
  \end{subfigure}
    \caption{Accuracy of \bert and \roberta on \pawsqqp decreases with increasing sentence length and parse tree height.
    }
  \label{fig:parser_height_length_pawsresult}
\end{figure*}

\begin{figure*}[h]
  \centering
  \begin{subfigure}[b]{0.49\linewidth}
    \includegraphics[width=\linewidth]{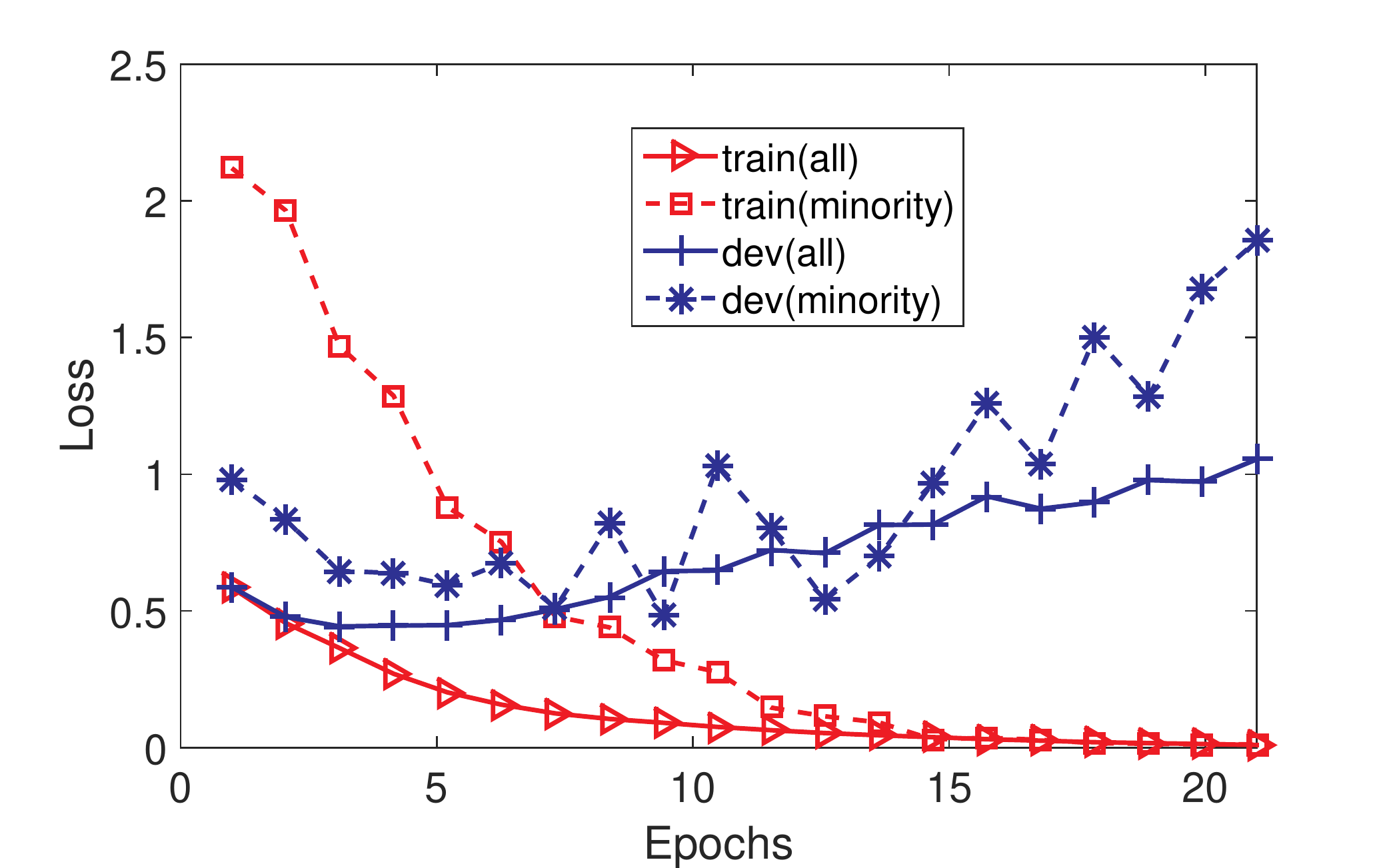}
    \caption{Loss}
  \end{subfigure}
  \begin{subfigure}[b]{0.49\linewidth}
    \includegraphics[width=\linewidth]{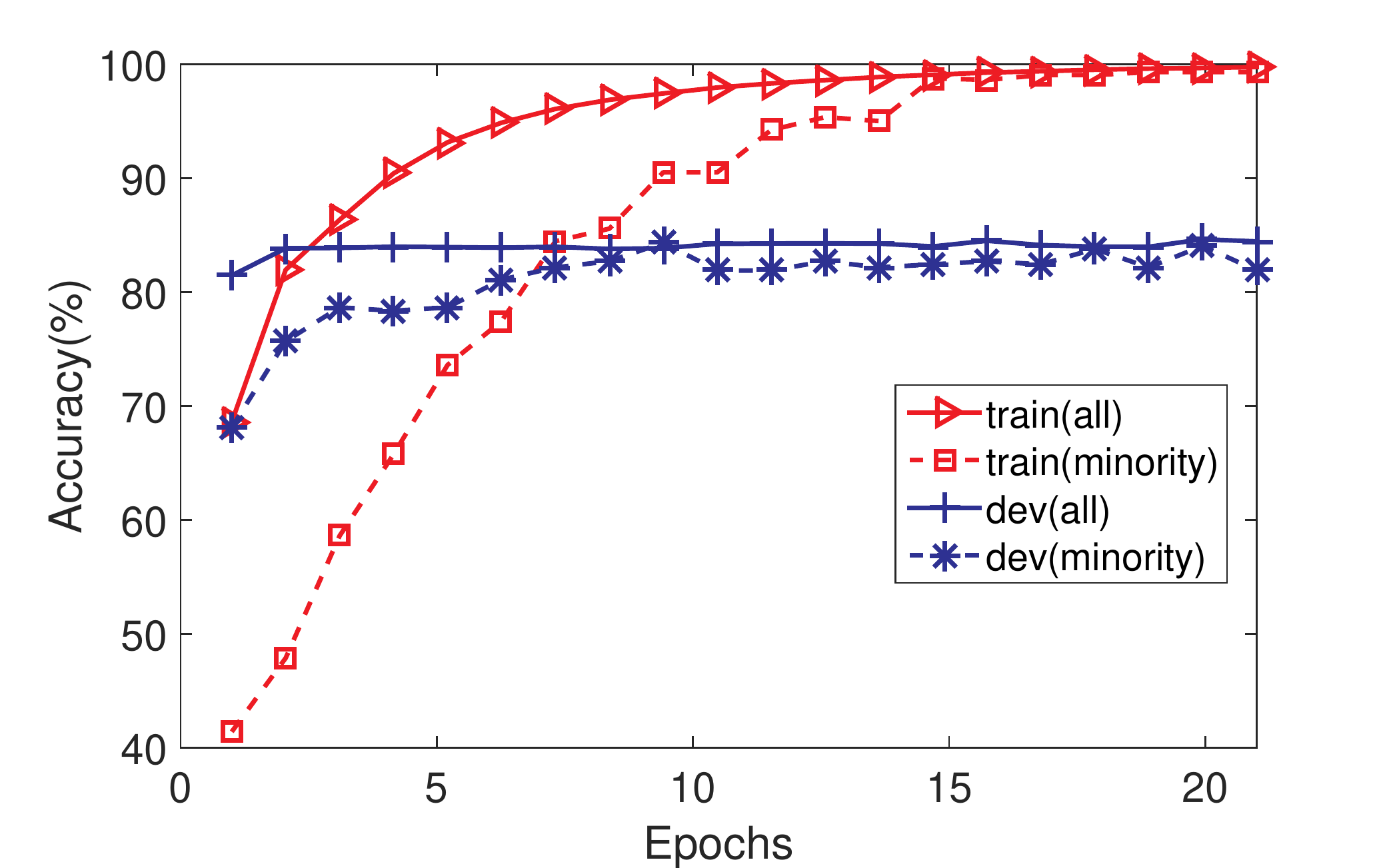}
    \caption{Accuracy}
  \end{subfigure}
  \caption{The average losses and accuracies of the examples in the training and dev set when fine-tuning \bert on MNLI.
    We show plots for the whole training set and the minority examples separately.
    The minority examples are non-entailment examples with at least 80\% premise-hypothesis overlap.
    Accuracy of minority examples takes longer to plateau.
    } 
  \label{fig:bertfinetuning-reason}
\end{figure*}

To test the hypothesis, we plot learning curves by fine-tuning pre-trained models on the challenging datasets directly \citep{liu2019inoculation}.
Specifically, we take 11,990 training examples from \pawsqqp,
and randomly sample the same number of training examples from \hans;\footnote{
    \hans has more examples in total (30,000), therefore
we sub-sample it to control for the data size.
}
the rest is used as dev/test set for evaluation.
In \reffig{learning_curve}, we see that all models reach 100\% accuracy rapidly on \hans.
However, on \paws, accuracy increases slowly and the models struggle to reach around 90\% accuracy even with the full training set.
This suggests that the amount of minority examples in QQP might not be sufficient for reliably estimating the model parameters.

To have a qualitative understanding on why \paws examples are difficult to learn,
we compare sentence length and constituency parse tree height of examples in \hans and \paws.\footnote{
    We use the off-the-shelf constituency parser from Stanford CoreNLP~\citep{manning-EtAl:2014:P14-5}.
    For each example, we compute the maximum length (number of words) and parse tree height of the two sentences.
} 
We find that \paws contains longer and syntactically more complex sentences,
with an average length of 20.7 words and parse tree height of 11.4,
compared to 9.2 and 7.5 on \hans.
\reffig{parser_height_length_pawsresult} shows that the accuracy of \bert and \roberta
on \pawsqqp
decreases as the example length and the parse tree height increase. 

\paragraph{Takeaway.}
We have shown that the inconsistent improvement on different challenging datasets
are resulted from the same mechanism:
pre-trained models improve robust accuracy by generalizing from minority examples,
however, perhaps unsurprisingly, different minority patterns may require varying amounts of training data.
This also poses a potential challenge in using data augmentation to tackle spurious correlations.

\subsection{Minority Examples Require Longer fine-tuning}
In the previos section, we have shown in \reffig{bert-finetuning} that longer fine-tuning improves accuracy on challenging examples,
even though the in-distribution accuracy saturates pretty quickly. 
To understand the result from the perspective of minority examples,
we compare the loss on all examples and the minority examples during fine-tuning.
\reffig{bertfinetuning-reason} shows the loss and accuracy at each epoch
on all examples and \hans-like examples in MNLI separately.

First, we see that the training loss of minority examples decreases more slowly than the average loss,
taking more than 15 epochs to reach near-zero loss.
Second, the dev accuracy curves show that the accuracy of minority examples plateaus later,
around epoch 10,
whereas the average accuracy stops to increaste around epoch 5.
In addition, it appears that BERT does not overfit with additional fine-tuning based on the accuracy curves.\footnote{
We find that the average accuracy stays almost the same while the dev loss is increasing.
\citet{guo2017calibration} had similar observations.
One possible explanation is that the model prediction becomes less confident (hence larger log loss), but the argmax prediction is correct.
}
Similary, a concurrent work \citep{zhang2020revisiting} has found that longer fine-tuning improve few-sample performance.

\paragraph{Takeaway.}
While longer fine-tuning does not help in-distribution accuracy,
we find that it improves performance on the minority groups.
This suggests that selecting models or early stopping based on the i.i.d. dev set performance is insufficient,
and we need new model selection criteria for robustness.

\section{Improve Generalization through Multi-task Learning}
Our results on minority examples 
show that increasing the amount of counterexamples to spurious correlations
helps to improve model robustness. 
Then, an obvious solution is data augmentation;
in fact, both \citet{mccoy2019hans} and \citet{zhang2019paws} show that adding a small amount of challenging examples to the training set
significantly improves performance on \hans and \paws.
However, these methods often require task-specific knowledge on spurious correlations
and heavy use of rules to generate the counterexamples. 
Instead of adding examples with specific patterns,
we investigate the effect of aggregating generic data from various sources through multi-task learning (MTL).
It has been shown that MTL reduces the sample complexity of individual tasks compared to single-task learning \cite{caruana1997multitask,baxter2000model,maurer2016benefit},
thus may further improve the generalization capability of pre-trained models, especially on the minority groups.

\begin{table*}[ht]
    \centering
\begin{tabular}{llccccc}
\toprule
             &            & \multicolumn{2}{c}{\textit{Task = MNLI}} & \multicolumn{3}{c}{\textit{Task = QQP}} \\
                            \cmidrule(lr){3-4} \cmidrule(lr){5-7}
             &            & In-distribution & Challenging & In-distribution & Challenging & Challenging \\
    \cent{Model} & Algo.  & \textbf{MNLI-m} & \textbf{HANS} & \textbf{QQP} & \textbf{\pawsqqp} & \textbf{\pawswiki}\\ 
\midrule
\multirow{2}{*}{\bert}    & STL   & 84.5 (0.1) & 62.5 (0.2) & 90.8 (0.3) & 36.1 (0.8) & 46.9 (0.3)\\         
                          & MTL   & 83.7 (0.3) & \textbf{68.2} (1.8) & 91.3 (.07) &  \textbf{45.9} (2.1) & \textbf{52.0} (1.9)\\  
\midrule
\multirow{2}{*}{\roberta} & STL   & 87.4 (0.2) & 74.1 (0.9) & 91.5 (0.2) & 42.6 (1.9)  & 49.6 (1.9) \\      
                          & MTL   & 86.4 (0.2) & 72.8 (2.4) & 91.7 (.04) & \textbf{51.7} (1.2)  & \textbf{57.7} (1.5)\\    
\bottomrule
\end{tabular}
    \caption{\label{tab:mtl_results}Comparison between models fine-tuned with multi-task (MTL) and single-task (STL) learning.
    MTL improves robust accuracy on challenging datasets.
    We ran $t$-tests for the mean accuracies of STL and MTL on five runs and the larger number is bolded when they differ significantly with a $p<0.001$.
    }
\end{table*}

\subsection{Multi-task Learning}
We learn from datasets from different sources jointly,
where one is the \textit{target dataset} to be evaluated on,
and the rest are \textit{auxiliary datasets}.
The target dataset and the auxiliary dataset can belong to either the same task, \eg MNLI and SNLI,
or different but related tasks, \eg MNLI and QQP.

All datasets share the representation given by the pre-trained model,
and we use separate linear classification layers for each dataset.
The learning objective is a weighted sum of average losses on each dataset.
We set the weight to be 1 for all datasets,
equivalent to sampling examples from each dataset proportional to its size.\footnote{
 Prior work has shown that the mixing weights may impact the final results in MTL,
 especially when there is a risk of overfitting to low-resource tasks \citep{raffel2019exploring}.
 Given the relatively large dataset sizes in our experiments (\reftab{aux_data}),
 we did not see significant change in the results when varying the mixing weights.
}
During training,
we sample mini-batches from each dataset sequentially 
and use the same optimization hyperparameters as in single-task fine-tuning (\refsec{finetuning})
except for smaller batch sizes due to memory constraints.\footnote{The minibatch size of the target dataset is 16. For the auxiliary dataset, it is proportional to the dataset size and not larger than 16, such that the total number of examples in a batch is at most 32.}.

\begin{table}[ht]
    \centering
    \begin{tabular}{lrcc}
        \toprule
        \multirow{2}{*}{Auxiliary Datasets} & \multirow{2}{*}{Size} & \multicolumn{2}{c}{Target} \\
        \cmidrule(lr){3-4}
                &      & NLI & PI \\
        \midrule
        MNLI & 393k & & $\checkmark$    \\
        SNLI & 549k & $\checkmark$ & $\checkmark$ \\
        QQP  & 364k & $\checkmark$ & \\
        \pawsall & 60k & $\checkmark$ & \\
        \hans & 30k &  & $\checkmark$ \\
        \bottomrule
    \end{tabular}
    \caption{\label{tab:aux_data}
    Auxiliary dataset sizes  for the different target datasets from two tasks: NLI and PI.
    }
\end{table}

\paragraph{Auxiliary datasets.}
We consider NLI and PI as related tasks since they both require understanding and comparing the meaning of two sentences. Therefore, we use both {benchmark datasets} and {challenging datasets} for NLI and PI as our auxiliary datasets. The hope is that benchmark data from related tasks helps transfer useful knowledge across tasks, thus improving generalization on minority examples, and the challenging datasets countering specific spurious correlations further improve generalization on the corresponding minority examples.
We analyze the contribution of the two types of auxiliary data in \refsec{mtl-results}.
The MTL training set up is shown in 
\reftab{aux_data}.\footnote{
    For MNLI, we did not include other PI datasets such as STS-B \citep{cer2017stsb} and MPRC \citep{dolan2005mrpc} since their sizes (3.7k and 7k) are too small compared to QQP and other auxiliary tasks.
}
Details on the auxiliary datasets are described in \refsec{datasets}.

\subsection{Results}
\label{sec:mtl-results}
\paragraph{MTL improves robust accuracy.}
Our main MTL results are shown in \reftab{mtl_results}.
MTL increases accuracies on the challenging datasets across tasks
without hurting the in-distribution performance,
especially when the minority examples in the target dataset is scarce (\eg \paws).
While prior work has shown limited success of MTL when tested on in-distribution data\citep{sogaard2016deep,hashimoto2017joint,raffel2019exploring},
our results demonstrate its value for out-of-distribution generalization. 

\begin{table}[ht]
\footnotesize{
    \centering
    \tabcolsep 4pt
    \begin{tabular}{llrrr}
    \toprule
        Model & Algo.  & \textbf{HANS-O}   & \textbf{HANS-C} & \textbf{HANS-S}     \\
      \midrule
        \bert   & STL  & 75.8 (4.9) & 59.1 (4.8) & 52.7 (1.2) \\
        \bert   & MTL  & 89.5 (1.9) & 61.9 (2.3) & 53.1 (1.1) \\
        \roberta & STL & 88.5 (2.0) & 70.0 (2.3) & 63.9 (1.4)  \\
        \roberta & MTL & 90.3 (1.2)  & 64.8 (3.1)  &63.5 (4.9)     \\
        \bottomrule
    \end{tabular}
    \caption{\label{tab:hans_category}
    MTL Results on different categories on \hans: \texttt{lexical overlap} (O), \texttt{constituent} (C), and \texttt{subsequence} (S).
    Both auxiliary data (MTL) and larger pre-training data (RoBERTa) improve accuracies mainly on \texttt{lexical overlap}.
    }
}
\end{table}

\begin{table}[ht]
\begin{center}
\begin{tabular}{lrrr}
\toprule
  Removed   &  \textbf{MNLI-m} & \textbf{HANS} & $\Delta$ \\ 
 \midrule

None
     &  83.7 (0.3) & 68.2 (1.8) & - \\  
\midrule
\pawsall            
      & 83.5 (0.3) & 64.6 (3.5) & -3.6\\   
QQP                               %
      & 83.2 (0.3) & 63.2 (3.7) & -5.0 \\ 
SNLI                              %
     & 84.3 (0.2) & 66.9 (1.5)  & -1.3 \\   
\bottomrule
\end{tabular}
    \caption{\label{tab:NLIablation}
    Results of the ablation study on auxiliary datasets using \bert on MNLI (the target task). 
    While the in-distribution performance is hardly affected when a specific auxiliary dataset is excluded,
    performance on the challenging data varies (difference shown in $\Delta$).
    }
\end{center}
\end{table}

On \hans, MTL improves the accuracy significantly for \bert but not for \roberta.
To confirm the result, we additionally experimented with \robertal and obtained consistent results:
MTL achieves an accuracy of 75.7 (2.1) on \hans,
similar to the STL result, 77.1 (1.6).
One potential explanation is that RoBERTa is already sufficient for providing good generalization from minority examples in MNLI. 

In addition, both MTL and \roberta yiedls biggest improvement on \texttt{lexical overlap},
as shown in the results on \hans by category (\reftab{hans_category}),
We believe the reason is that \texttt{lexical overlap} is the most representative pattern among high-overlap and non-entailment training examples.
In fact, 
85\% of the 727 \hans-like examples belongs to \texttt{lexical overlap}.
This suggests that further improvement on \hans may require better data coverage on other categories.

On \paws, MTL consistently yields large improvement across pre-trained models.
Given that QQP has fewer minority examples resembling the patterns in \paws,
which is also harder to learn (\refsec{paws}),
the results show that MTL is an effective way to improve generalization when the minority examples are scarce.
Next, we investigate why MTL is helpful.

\begin{table}[t]
\begin{center}
\begin{tabular}{lrrr}
\toprule
  Removed   &  \textbf{QQP} & \textbf{\pawsqqp} & $\Delta$ \\ 
 \midrule

None
     &  91.3 (.07) & 45.9 (2.1) & - \\  
\midrule
HANS           
      & 91.5 (.06) & 45.3 (1.8) & -0.6 \\   
MNLI                               %
      & 91.2 (.11) & 42.3 (1.8) & -3.6 \\ 
SNLI                              %
     & 91.3 (.09) & 44.2 (1.3)  & -1.7 \\   
\bottomrule
\end{tabular}
    \caption{\label{tab:PIablation}
    Results of the ablation study on auxiliary datasets using \bert on QQP (the target task). 
    While the in-distribution performance is hardly affected when a specific auxiliary dataset is excluded,
    performance on the challenging data varies (difference shown in $\Delta$). 
    }
\end{center}
\end{table}

\paragraph{Improved generalization from minority examples.}
We are interested in finding how MTL helps generalization from minority examples.
One possible explanation is that the challenging data in the auxiliary datasets 
prevent the model from learning suprious patterns.
However, the ablation studies on auxiliary datasets in \reftab{NLIablation} and \reftab{PIablation}
show that the challenging datasets are not much more helpful than benchmark datasets.
The other possible explanation is that MTL reduces sample complexity for learning from the minority examples in the \emph{target dataset}.
To verify this, we remove minority examples from both the auxiliary and the target datasets,
and compare their effect on the robust accuracy.

We focus on PI because MTL shows largest improvement there.
In Table~\ref{tab:MTLablation}, we show the results after removing minority examples in the target dataset, QQP, and the auxiliary dataset, MNLI, respectively.
We also add a control baseline where the same amounts of randomly sampled examples are removed.
The results confirm our hypothesis:
without the minority examples in the target dataset,
MTL is only marginally better than STL on \pawsqqp.
In contrast, 
removing minority examples in the auxiliary dataset has a similar effect to removing random examples;
both do not cause significant performance drop.
Therefore, we conclude that MTL improves robust accuracy by improving generalization from minority examples in the target dataset.

\paragraph{Takeaway.}
These results suggest that
both pre-training and MTL do not enable extrapolation,
instead, they improve generalization from minority examples in the (target) training set.
Thus it is important to increase coverage of diverse patterns in the data to improve robustness to spurious correlations.

\begin{table}[ht]
\begin{center}
\begin{tabular}{lrrr}
\toprule
  Removed   &  \textbf{QQP} & \textbf{\pawsqqp} & $\Delta$ \\ 
 \midrule

None
     &  91.3 (.07) & 45.9 (2.1) & - \\  
\midrule

\multicolumn{4}{l}{\textit{random examples}} \\
QQP           
      & 91.3 (.03) & 44.3 (.31 ) &  -1.6\\    
MNLI                             %
      & 91.4 (.02) & 45.0 (1.5 ) &  -0.9\\ 

\multicolumn{4}{l}{\textit{minority examples}} \\
QQP           
      & 91.3 (.09) & \textbf{38.2} (.73) & -7.7 \\    
MNLI                             %
      & 91.3 (.08) & 44.3 (2.0) & -1.6 \\ 
\bottomrule
\end{tabular}
    \caption{\label{tab:MTLablation}
    Ablation study on the effect of minority examples in the auxiliary (MNLI) and the target (QQP) datasets in MTL with \bert.
    For MNLI, we removed 727 non-entailment examples with 100\% overlap.
    For QQP, we removed 228 non-paraphrase examples with 100\% overlap.
    We also removed equal amounts of random examples in the control experiments.
    We ran $t$-tests for the mean accuracies after minority removal and random removal based on five runs, and numbers with a significant difference ($p<0.001$) are bolded.
    The improvement from MTL mainly comes from better generalization from minority examples in the \emph{target dataset}.
    }
\end{center}
\end{table}

\section{Related Work}
\label{sec:related}

\paragraph{Pre-training and robustness.}
Recently, there is an increasing amount of interest in studying the effect of pre-training on robustness. 
\citet{hendrycks2019pretraining,hendrycks2020pretrained} show that pre-training improves model robustness to label noise, class imbalance, and out-of-distribution detection. 
In cross-domain question-answering, \citet{dnetli2019} show that the ensemble of different pre-trained models significantly improves performance on out-of-domain data.
In this work, we answers \emph{why} pre-trained models appear to improve out-of-distribution robustness
and point out the importance of minority examples in the training data.

\paragraph{Data augmentation.}
The most straightforward way to improve model robustness to out-of-distribution data is
to augment the training set with examples from the target distribution.
Recent work has shown that augmenting syntactically-rich examples improves robust accuracy on NLI \citep{min2020syntactic}.
Similarly, counterfactual augmentation aims to identify parts of the input that impact the label when intervened upon,
thus avoiding learning spurious features \citep{goyal2019counterfactual,kaushik2020learning}. 
Finally, data recombination has been used to achieve compositional generalization
\citep{jia2016recombination,andreas2020compositional}.
However, data augmentation techniques largely rely on prior knowledge of the spurious correlations or human efforts. 
In addition, as shown in \refsec{paws} and a concurrent work \citep{jha2020acl},
it is often unclear how much augmented data is needed for learning a pattern.
Our work shows promise in adding generic pre-training data or related auxiliary data (through MTL)
without assumptions on the target distribution.

\paragraph{Robust learning algorithms.}
Serveral recent work proposes new learning algorithms that are robust to spurious correlations in NLI datasets \citep{he2019unlearn,clark2019ensemble,yaghoobzadeh2019,zhou2020towards,sagawa2020distributionally,mahabadi2020end,utama-etal-2020-mind}.
They rely on prior knowledge 
to focus on ``harder'' examples that do not enable shortcuts during training.
One weakness of these methods is their arguably strong assumption on knowing the spurious correlations \emph{a priori}. 
Our work provides evidence that large amounts of \emph{generic} data can be used to improve out-of-distribution generalization.
Similarly, recent work has shown that semi-supervised learning with generic auxiliary data improves model robustness to adversarial examples \citep{schmidt2018adversarially,carmon2019unlabeled}.

\paragraph{Transfer learning.}
Robust learning is also related to domain adaptation or transfer learning
since both aim to learn from one distribution and achieve good performance on a different but related target distribution.
Data selection and reweighting are common techniques used in domain adaptation.
Similar to our findings on minority examples,
source examples similar to the target data have been found to be helpful to transfer \citep{ruder2017learning,liu2019reinforced}.
In addition, many works have shown that 
MTL improves model performance on out-of-domain datasets \citep{ruder2017overview,dnetli2019,liu-etal-2019-multi}.
A concurrent work \citep{akula2020words} shows that MTL improves robustness on advesarial examples in visual grounding.
In this work, we further 
connect the effectiveness of MTL to generalization from minority examples.

\section{Conclusion and Discussion}
Our study is motivated by recent observations on the robustness of large-scale pre-trained transformers. 
Specifically, we focus on robust accuracy on the challenging datasets
which are designed to expose spurious correlations learned by the model.
Our analysis reveals that pre-training improves robustness by better generalizing from a minority of examples that counter dominant spurious patterns in the training set.
In addition, we show that more pre-training data, larger model size, and additional auxiliary data through MTL 
further improve robustness, especially when the amount of minority examples is scarce.

Our work suggests that it is possible to go beyond the robustness-accuracy trade-off with more data.
However, the amount of improvement is still limited by the coverage of the training data
because current models do not extrapolate to unseen patterns.
Thus an important future direction is to increase data diversity through new crowdsourcing protocols or efficient human-in-the-loop augmentation.

While our work provides new perspectives on pre-training and robustness,
it only scratches the surface of the effectiveness of pre-trained models
and leaves many questions open.
For example, why pre-trained models do not overfit to the minority examples;
how different initialization (from different pre-trained models) influences optimization and generalization.
Understanding these questions are key to designing better pre-training methods for robust models. 

Finally, the difference between results on \hans and \paws calls for more careful thinking on the formulation and evaluation of out-of-distribution generalization.
Semi-manually constructed challenging data often covers only a specific type of distribution shift, 
thus the results may not generalize to other types.
A more comprehensive evaluation will drive the development of principled methods for out-of-distribution generalization.

\section*{Acknowledgments}
We would like to thank the Lex and Comprehend groups at Amazon Web Services AI for helpful discussions, and the reviewers for their insightful comments. We would also like to thank the GluonNLP team for the infrastructure support.

\bibliography{tacl2018,all}

\begin{thebibliography}{50}
\expandafter\ifx\csname natexlab\endcsname\relax\def\natexlab#1{#1}\fi

\bibitem[{Akula et~al.(2020)Akula, Gella, Al-Onaizan, Zhu, and
  Reddy}]{akula2020words}
Arjun~R Akula, Spandana Gella, Yaser Al-Onaizan, Song-Chun Zhu, and Siva Reddy.
  2020.
\newblock Words aren't enough, their order matters: On the robustness of
  grounding visual referring expressions.
\newblock In \emph{Association for Computational Linguistics (ACL)}.

\bibitem[{Andreas(2020)}]{andreas2020compositional}
J.~Andreas. 2020.
\newblock Good-enough compositional data augmentation.
\newblock In \emph{Association for Computational Linguistics (ACL)}.

\bibitem[{Arjovsky et~al.(2019)Arjovsky, Bottou, Gulrajani, and
  Lopez-Paz}]{arjovsky2019invariant}
M.~Arjovsky, L.~Bottou, I.~Gulrajani, and D.~Lopez-Paz. 2019.
\newblock Invariant risk minimization.
\newblock \emph{arXiv preprint arXiv:1907.02893v2}.

\bibitem[{Baxter(2000)}]{baxter2000model}
J.~Baxter. 2000.
\newblock A model of inductive bias learning.
\newblock \emph{Journal of Artificial Intelligence Research (JAIR)},
  12:149--198.

\bibitem[{Bowman et~al.(2015)Bowman, Angeli, Potts, and
  Manning}]{bowman2015large}
S.~Bowman, G.~Angeli, C.~Potts, and C.~D. Manning. 2015.
\newblock A large annotated corpus for learning natural language inference.
\newblock In \emph{Empirical Methods in Natural Language Processing (EMNLP)}.

\bibitem[{Carmon et~al.(2019)Carmon, Raghunathan, Schmidt, Liang, and
  Duchi}]{carmon2019unlabeled}
Y.~Carmon, A.~Raghunathan, L.~Schmidt, P.~Liang, and J.~C. Duchi. 2019.
\newblock Unlabeled data improves adversarial robustness.
\newblock In \emph{Advances in Neural Information Processing Systems
  (NeurIPS)}.

\bibitem[{Caruana(1997)}]{caruana1997multitask}
Rich Caruana. 1997.
\newblock Multitask learning.
\newblock \emph{Machine learning}, 28(1):41--75.

\bibitem[{Cer et~al.(2017)Cer, Diab, Agirre, Lopez-Gazpio, and
  Specia}]{cer2017stsb}
D.~Cer, M.~Diab, E.~Agirre, I.~Lopez-Gazpio, and L.~Specia. 2017.
\newblock {S}em{E}val-2017 task 1: Semantic textual similarity - multilingual
  and cross-lingual focused evaluation.
\newblock In \emph{Proceedings of the Eleventh International Workshop on
  Semantic Evaluations}.

\bibitem[{Clark et~al.(2019)Clark, Yatskar, and
  Zettlemoyer}]{clark2019ensemble}
C.~Clark, M.~Yatskar, and L.~Zettlemoyer. 2019.
\newblock Don't take the easy way out: Ensemble based methods for avoiding
  known dataset biases.
\newblock In \emph{Empirical Methods in Natural Language Processing (EMNLP)}.

\bibitem[{Dasgupta et~al.(2018)Dasgupta, Guo, Stuhlm{\"{u}}ller, Gershman, and
  Goodman}]{DasguptaGSGG18}
Ishita Dasgupta, Demi Guo, Andreas Stuhlm{\"{u}}ller, Samuel Gershman, and Noah
  Goodman. 2018.
\newblock Evaluating compositionality in sentence embeddings.
\newblock In \emph{Annual Meeting of the Cognitive Science Society, CogSci
  2018}.

\bibitem[{Devlin et~al.(2019)Devlin, Chang, Lee, and
  Toutanova}]{devlin2019BERT}
J.~Devlin, M.~Chang, K.~Lee, and K.~Toutanova. 2019.
\newblock Bert: Pre-training of deep bidirectional transformers for language
  understanding.
\newblock In \emph{North American Association for Computational Linguistics
  (NAACL)}.

\bibitem[{Dolan and Brockett(2005)}]{dolan2005mrpc}
W.~B. Dolan and C.~Brockett. 2005.
\newblock Automatically constructing a corpus of sentential paraphrases.
\newblock In \emph{Proceedings of the International Workshop on Paraphrasing}.

\bibitem[{Glockner et~al.(2018)Glockner, Shwartz, and
  Goldberg}]{glockner2018breaking}
M.~Glockner, V.~Shwartz, and Y.~Goldberg. 2018.
\newblock Breaking {NLI} systems with sentences that require simple lexical
  inferences.
\newblock In \emph{Association for Computational Linguistics (ACL)}.

\bibitem[{Goyal et~al.(2019)Goyal, Wu, Ernst, Batra, Parikh, and
  Lee}]{goyal2019counterfactual}
Y.~Goyal, Z.~Wu, J.~Ernst, D.~Batra, D.~Parikh, and S.~Lee. 2019.
\newblock Counterfactual visual explanations.
\newblock In \emph{International Conference on Machine Learning (ICML)}.

\bibitem[{Guo et~al.(2017)Guo, Pleiss, Sun, and
  Weinberger}]{guo2017calibration}
C.~Guo, G.~Pleiss, Y.~Sun, and K.~Q. Weinberger. 2017.
\newblock On calibration of modern neural networks.
\newblock In \emph{International Conference on Machine Learning (ICML)}.

\bibitem[{Guo et~al.(2020)Guo, He, He, Lausen, Li, Lin, Shi, Wang, Xie, Zha,
  Zhang, Zhang, Zhang, Zhang, Zheng, and Zhu}]{guo2020gluoncv}
J.~Guo, H.~He, T.~He, L.~Lausen, M.~Li, H.~Lin, X.~Shi, C.~Wang, J.~Xie,
  S.~Zha, A.~Zhang, H.~Zhang, Z.~Zhang, Z.~Zhang, S.~Zheng, and Y.~Zhu. 2020.
\newblock Gluoncv and gluonnlp: Deep learning in computer vision and natural
  language processing.
\newblock \emph{Journal of Machine Learning Research (JMLR)}, 21.

\bibitem[{Gururangan et~al.(2018)Gururangan, Swayamdipta, Levy, Schwartz,
  Bowman, and Smith}]{gururangan2018annotation}
S.~Gururangan, S.~Swayamdipta, O.~Levy, R.~Schwartz, S.~R. Bowman, and N.~A.
  Smith. 2018.
\newblock Annotation artifacts in natural language inference data.
\newblock In \emph{North American Association for Computational Linguistics
  (NAACL)}.

\bibitem[{Hashimoto et~al.(2017)Hashimoto, Xiong, Tsuruoka, and
  Socher}]{hashimoto2017joint}
K.~Hashimoto, C.~Xiong, Y.~Tsuruoka, and R.~Socher. 2017.
\newblock A joint many-task model: Growing a neural network for multiple {NLP}
  tasks.
\newblock In \emph{Empirical Methods in Natural Language Processing (EMNLP)}.

\bibitem[{He et~al.(2019)He, Zha, and Wang}]{he2019unlearn}
H.~He, S.~Zha, and H.~Wang. 2019.
\newblock Unlearn dataset bias for natural language inference by fitting the
  residual.
\newblock In \emph{Proceedings of the EMNLP Workshop on Deep Learning for
  Low-Resource {NLP}}.

\bibitem[{Hendrycks et~al.(2019)Hendrycks, Lee, and
  Mazeika}]{hendrycks2019pretraining}
D.~Hendrycks, K.~Lee, and M.~Mazeika. 2019.
\newblock Using pre-training can improve model robustness and uncertainty.
\newblock In \emph{International Conference on Machine Learning (ICML)}.

\bibitem[{Hendrycks et~al.(2020)Hendrycks, Liu, Wallace, Dziedzic, Krishnan,
  and Song}]{hendrycks2020pretrained}
D.~Hendrycks, X.~Liu, E.~Wallace, A.~Dziedzic, R.~Krishnan, and D.~Song. 2020.
\newblock Pretrained transformers improve out-of-distribution robustness.
\newblock In \emph{Association for Computational Linguistics (ACL)}.

\bibitem[{Iyer et~al.(2017)Iyer, Dandekar, and Csernai}]{iyer2017qqp}
S.~Iyer, N.~Dandekar, and K.~Csernai. 2017.
\newblock First quora dataset release: Question pairs.
\newblock \emph{Accessed online at
  https://www.quora.com/q/quoradata/First-Quora-Dataset-Release-Question-Pairs}.

\bibitem[{Jha et~al.(2020)Jha, Lovering, and Pavlick}]{jha2020acl}
R.~Jha, C.~Lovering, and E.~Pavlick. 2020.
\newblock When does data augmentation help generalization in {NLP}?
\newblock In \emph{Association for Computational Linguistics (ACL)}.

\bibitem[{Jia and Liang(2016)}]{jia2016recombination}
R.~Jia and P.~Liang. 2016.
\newblock Data recombination for neural semantic parsing.
\newblock In \emph{Association for Computational Linguistics (ACL)}.

\bibitem[{Kaushik et~al.(2020)Kaushik, Hovy, and Lipton}]{kaushik2020learning}
D.~Kaushik, E.~Hovy, and Z.~C. Lipton. 2020.
\newblock Learning the difference that makes a difference with
  counterfactually-augmented data.
\newblock In \emph{International Conference on Learning Representations
  (ICLR)}.

\bibitem[{Li et~al.(2019)Li, Zhang, Liu, Zhang, Wang, Zhou, Liu, Wu, and
  Wang}]{dnetli2019}
Hongyu Li, Xiyuan Zhang, Yibing Liu, Yiming Zhang, Quan Wang, Xiangyang Zhou,
  Jing Liu, Hua Wu, and Haifeng Wang. 2019.
\newblock D-net: A pre-training and fine-tuning framework for improving the
  generalization of machine reading comprehension.
\newblock In \emph{Proceedings of the 2nd Workshop on Machine Reading for
  Question Answering}, pages 212--219.

\bibitem[{Liu et~al.(2019{\natexlab{a}})Liu, Song, Zou, and
  Zhang}]{liu2019reinforced}
Miaofeng Liu, Yan Song, Hongbin Zou, and Tong Zhang. 2019{\natexlab{a}}.
\newblock Reinforced training data selection for domain adaptation.
\newblock In \emph{Proceedings of the 57th Annual Meeting of the Association
  for Computational Linguistics}, pages 1957--1968.

\bibitem[{Liu et~al.(2019{\natexlab{b}})Liu, Schwartz, and
  Smith}]{liu2019inoculation}
N.~F. Liu, R.~Schwartz, and N.~A. Smith. 2019{\natexlab{b}}.
\newblock Inoculation by fine-tuning: A method for analyzing challenge
  datasets.
\newblock In \emph{North American Association for Computational Linguistics
  (NAACL)}.

\bibitem[{Liu et~al.(2019{\natexlab{c}})Liu, He, Chen, and
  Gao}]{liu-etal-2019-multi}
Xiaodong Liu, Pengcheng He, Weizhu Chen, and Jianfeng Gao. 2019{\natexlab{c}}.
\newblock Multi-task deep neural networks for natural language understanding.
\newblock In \emph{Association for Computational Linguistics (ACL)}.

\bibitem[{Liu et~al.(2019{\natexlab{d}})Liu, Ott, Goyal, Du, Joshi, Chen, Levy,
  Lewis, Zettlemoyer, and Stoyanov}]{liu2019roberta}
Y.~Liu, M.~Ott, N.~Goyal, J.~Du, M.~Joshi, D.~Chen, O.~Levy, M.~Lewis,
  L.~Zettlemoyer, and V.~Stoyanov. 2019{\natexlab{d}}.
\newblock {R}o{BERT}a: A robustly optimized {BERT} pretraining approach.
\newblock \emph{arXiv preprint arXiv:1907.11692}.

\bibitem[{Mahabadi et~al.(2020)Mahabadi, Belinkov, and
  Henderson}]{mahabadi2020end}
R.~K. Mahabadi, Y.~Belinkov, and J.~Henderson. 2020.
\newblock End-to-end bias mitigation by modelling biases in corpora.
\newblock In \emph{Association for Computational Linguistics (ACL)}.

\bibitem[{Manning et~al.(2014)Manning, Surdeanu, Bauer, Finkel, Bethard, and
  McClosky}]{manning-EtAl:2014:P14-5}
Christopher~D. Manning, Mihai Surdeanu, John Bauer, Jenny Finkel, Steven~J.
  Bethard, and David McClosky. 2014.
\newblock The {Stanford} {CoreNLP} natural language processing toolkit.
\newblock In \emph{Association for Computational Linguistics (ACL) System
  Demonstrations}.

\bibitem[{Maurer et~al.(2016)Maurer, Pontil, and
  Romera-Paredes}]{maurer2016benefit}
A.~Maurer, M.~Pontil, and B.~Romera-Paredes. 2016.
\newblock The benefit of multitask representation learning.
\newblock \emph{Journal of Machine Learning Research (JMLR)}, 17:1--32.

\bibitem[{McCoy et~al.(2019)McCoy, Pavlick, and Linzen}]{mccoy2019hans}
R.~T. McCoy, E.~Pavlick, and T.~Linzen. 2019.
\newblock Right for the wrong reasons: Diagnosing syntactic heuristics in
  natural language inference.
\newblock \emph{arXiv preprint arXiv:1902.01007}.

\bibitem[{Min et~al.(2020)Min, McCoy, Das, Pitler, and
  Linzen}]{min2020syntactic}
J.~Min, R.~T. McCoy, D.~Das, E.~Pitler, and T.~Linzen. 2020.
\newblock Syntactic data augmentation increases robustness to inference
  heuristics.
\newblock In \emph{Association for Computational Linguistics (ACL)}.

\bibitem[{Nie et~al.(2019)Nie, Wang, and Bansal}]{nie2019analyzing}
Yixin Nie, Yicheng Wang, and Mohit Bansal. 2019.
\newblock Analyzing compositionality-sensitivity of nli models.
\newblock In \emph{Proceedings of the AAAI Conference on Artificial
  Intelligence}, volume~33, pages 6867--6874.

\bibitem[{Nye et~al.(2019)Nye, Solar-Lezama, Tenenbaum, and
  Lake}]{nye2019learning}
M.~I. Nye, A.~Solar-Lezama, J.~B. Tenenbaum, and B.~M. Lake. 2019.
\newblock Learning compositional rules via neural program synthesis.
\newblock In \emph{Advances in Neural Information Processing Systems
  (NeurIPS)}.

\bibitem[{Oren et~al.(2019)Oren, Sagawa, Hashimoto, and
  Liang}]{oren2019distributionally}
Y.~Oren, S.~Sagawa, T.~B. Hashimoto, and P.~Liang. 2019.
\newblock Distributionally robust language modeling.
\newblock In \emph{Empirical Methods in Natural Language Processing (EMNLP)}.

\bibitem[{Raffel et~al.(2019)Raffel, Shazeer, Roberts, Lee, Narang, Matena,
  Zhou, Li, and Liu}]{raffel2019exploring}
C.~Raffel, N.~Shazeer, A.~Roberts, K.~Lee, S.~Narang, M.~Matena, Y.~Zhou,
  W.~Li, and P.~J. Liu. 2019.
\newblock Exploring the limits of transfer learning with a unified text-to-text
  transformer.
\newblock \emph{arXiv preprint arXiv:1910.10683}.

\bibitem[{Ruder(2017)}]{ruder2017overview}
S.~Ruder. 2017.
\newblock An overview of multi-task learning in deep neural networks.
\newblock \emph{arXiv preprint arXiv:1706.05098}.

\bibitem[{Ruder and Plank(2017)}]{ruder2017learning}
Sebastian Ruder and Barbara Plank. 2017.
\newblock Learning to select data for transfer learning with bayesian
  optimization.
\newblock \emph{arXiv preprint arXiv:1707.05246}.

\bibitem[{Sagawa et~al.(2020)Sagawa, Koh, Hashimoto, and
  Liang}]{sagawa2020distributionally}
S.~Sagawa, P.~W. Koh, T.~B. Hashimoto, and P.~Liang. 2020.
\newblock Distributionally robust neural networks for group shifts: On the
  importance of regularization for worst-case generalization.
\newblock In \emph{International Conference on Learning Representations
  (ICLR)}.

\bibitem[{Schmidt et~al.(2018)Schmidt, Santurkar, Tsipras, Talwar, and
  Madry}]{schmidt2018adversarially}
L.~Schmidt, S.~Santurkar, D.~Tsipras, K.~Talwar, and A.~Madry. 2018.
\newblock Adversarially robust generalization requires more data.
\newblock In \emph{Advances in Neural Information Processing Systems
  (NeurIPS)}, pages 5014--5026.

\bibitem[{S{\o}gaard and Goldberg(2016)}]{sogaard2016deep}
A.~S{\o}gaard and Y.~Goldberg. 2016.
\newblock Deep multi-task learning with low level tasks supervised at lower
  layers.
\newblock In \emph{Association for Computational Linguistics (ACL)}.

\bibitem[{Utama et~al.(2020)Utama, Moosavi, and
  Gurevych}]{utama-etal-2020-mind}
Prasetya~Ajie Utama, Nafise~Sadat Moosavi, and Iryna Gurevych. 2020.
\newblock \href {https://doi.org/10.18653/v1/2020.acl-main.770} {Mind the
  trade-off: Debiasing {NLU} models without degrading the in-distribution
  performance}.
\newblock In \emph{Proceedings of the 58th Annual Meeting of the Association
  for Computational Linguistics}, pages 8717--8729, Online. Association for
  Computational Linguistics.

\bibitem[{Williams et~al.(2017)Williams, Nangia, and
  Bowman}]{williams2017broad}
A.~Williams, N.~Nangia, and S.~R. Bowman. 2017.
\newblock A broad-coverage challenge corpus for sentence understanding through
  inference.
\newblock \emph{arXiv preprint arXiv:1704.05426}.

\bibitem[{Yaghoobzadeh et~al.(2019)Yaghoobzadeh, des Combes, Hazen, and
  Sordoni}]{yaghoobzadeh2019}
Yadollah Yaghoobzadeh, Remi~Tachet des Combes, Timothy~J. Hazen, and Alessandro
  Sordoni. 2019.
\newblock Robust natural language inference models with example forgetting.
\newblock \emph{CoRR}, abs/1911.03861.

\bibitem[{Zhang et~al.(2020)Zhang, Wu, Katiyar, Weinberger, and
  Artzi}]{zhang2020revisiting}
T.~Zhang, F.~Wu, A.~Katiyar, K.~Q. Weinberger, and Y.~Artzi. 2020.
\newblock Revisiting few-sample {BERT} fine-tuning.
\newblock \emph{arXiv preprint arXiv:2006.05987}.

\bibitem[{Zhang et~al.(2019)Zhang, Baldridge, and He}]{zhang2019paws}
Y.~Zhang, J.~Baldridge, and L.~He. 2019.
\newblock {PAWS}: Paraphrase adversaries from word scrambling.
\newblock In \emph{North American Association for Computational Linguistics
  (NAACL)}.

\bibitem[{Zhou and Bansal(2020)}]{zhou2020towards}
X.~Zhou and M.~Bansal. 2020.
\newblock Towards robustifying {NLI} models against lexical dataset biases.
\newblock In \emph{Association for Computational Linguistics (ACL)}.

\end{thebibliography}
\bibliographystyle{acl_natbib}

\end{document}